\documentclass{article}
\usepackage{spconf}
\usepackage{graphicx}%
\usepackage{bigstrut,multirow,rotating,booktabs}
\usepackage{amsmath,amssymb,amsfonts}%
\pdfoutput=1

\title{Gabor-guided transformer for single image deraining}
\name{Sijin He,Guangfeng Lin*\thanks{*corresponding author}}
\address{Xi'an University of Technology,China}
\begin{document}
%\ninept
%
\topmargin=0mm
\maketitle
\small
\begin{abstract}
	\label{sec:intro}
Image deraining  have  have gained a great deal of attention in order to address the challenges posed by the effects of harsh weather conditions on visual tasks. While convolutional neural networks (CNNs) are popular, their limitations in capturing global information may result in ineffective rain removal. Transformer-based methods with self-attention mechanisms have improved, but they tend to distort high-frequency details that are crucial for image fidelity. To solve this problem, we propose the Gabor-guided tranformer (Gabformer) for single image deraining. The focus on local texture features is enhanced by incorporating the information processed by the Gabor filter into the query vector, which also improves the robustness of the model to noise due to the properties of the filter. Extensive experiments on the benchmarks demonstrate that our method outperforms state-of-the-art approaches.
\end{abstract}

\section{Introduction}
\label{sec:intro}
Images captured in adverse rainy conditions can significantly affect the performance of high-level vision tasks. To overcome this challenge, various techniques have been developed to reduce the negative impact of rain on images\cite{1,2,3}. Among these methods, convolutional neural networks (CNNs) have received a lot of attention for their excellent ability to learn complex hierarchical features from data.

Many deep learning approaches concentrate on enhancing CNN frameworks, considering them a more desirable option\cite{4,5,6,7,8}. However, the convolutional operation of CNNs has limitations in processing global information, which makes it difficult to capture long-range correlation between pixels. 
\begin{figure}[htb]
	\begin{minipage}[b]{1.0\linewidth}
		\centering
		\centerline{\includegraphics[width=1\textwidth]{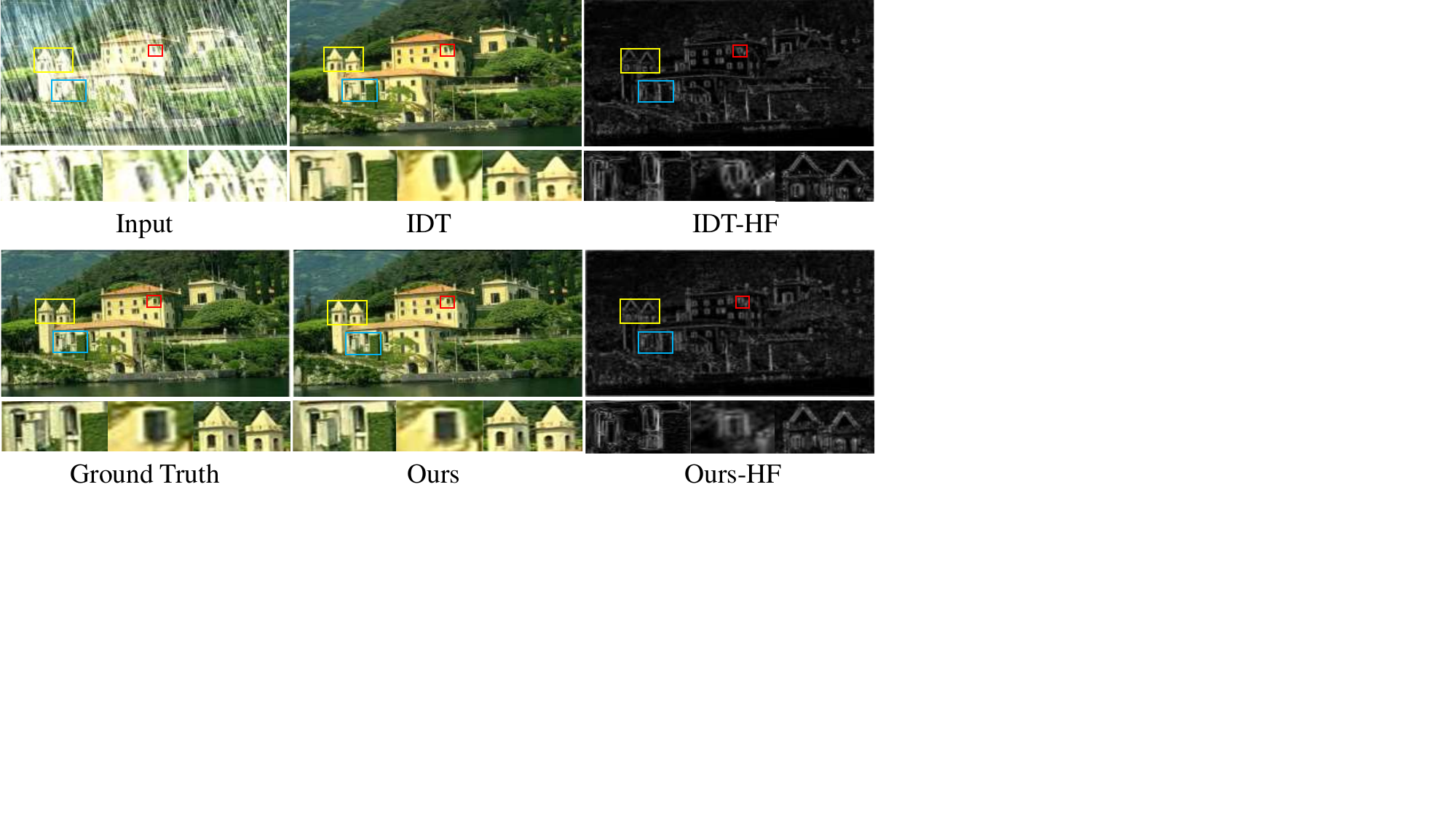}}
		%  \vspace{2.0cm}
	\end{minipage}
	\caption{Visual comparison effect of our method with IDT\cite{2022IDT}. HF is the visualization result after high-pass filtering. Unlike the IDT, which adds priors to the network structure, our method extract the texture through Gabor, which can recover a finer effect.}
	\label{fig:res}
\end{figure}
In recent years, transformer-based methods have made significant progress in the field of image deraining \cite{10,11,12}. The self-attention mechanism\cite{9} captures global information more efficiently and improves image distortion caused by rain. Rain-induced image distortion typically involves minute details and textures that are crucial for restoring image clarity and realism. However, we observe that the self-attention mechanism favors focusing on the global information in the image and is relatively poor at processing some high-frequency information (e.g., subtle image textures), as shown in Fig. 1. Therefore, providing the network with relevant high-frequency information can enhance its recovery ability. The standard transformer\cite{9} typically uses a global relationship of query-key pairs to aggregate image features. Both queries and keys are derived by mapping the same input without any processing to generate them. The query measures the importance of different locations in the image and it plays a guiding role in computing attention, determining which parts of the input image the model focuses on. The transformer's performance can be improved through the provision of queries with effective features.

In this study, we propose a new Gabor-guided transformer  for image deraining to solve the above problem.
Gabor filter are known for its selectivity in multiple scales, directions and frequencies, as well as for its remarkable effect on texture extraction.This flexibility is able to sensitively capture the texture information in the image at different scales and directions, and show robustness to illumination changes\cite{13}.
We provide more local texture information to the queries of the self-attention mechanism by using a modified Gabor filter, which tends to focus on regions associated with high-frequency details of the image in the attention computation to enhance attention to local features. In addition, the local texture information extracted by the Gabor filter is usually robust to noise and image changes. By introducing this robustness into the query vector, the self-attention mechanism is better able to resist the effects of noise or subtle changes when computing the attention weights, thus improving the robustness of the model, enabling it to better handle image data from complex environments, and improving the model's adaptability to imperfect conditions that may occur in real scenes. We apply the cross-channel self-attention mechanism to reduce the computational complexity of the model\cite{zamir2022restormer}. Our method outperforms state-of-the-art methods in extensive benchmark experiments.
\begin{figure*}[htb]
	\begin{minipage}[b]{1.0\linewidth}
		\centering
		\centerline{\includegraphics[width=0.95\textwidth,height=0.55\textwidth]{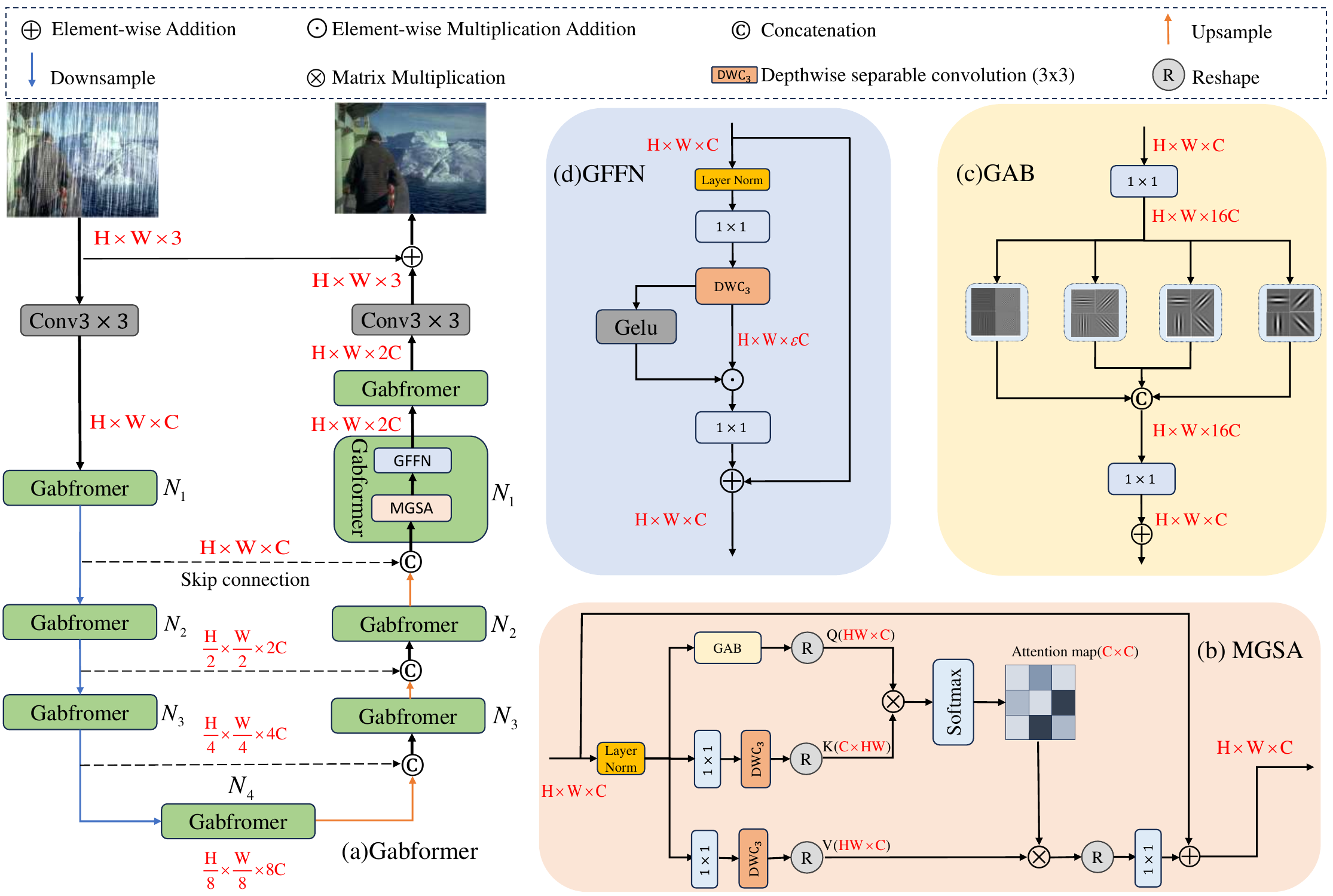}}
		%  \vspace{2.0cm}
	\end{minipage}
	\caption{Detailed framework of Gabformer with the main constituent modules of 
		(a) overall framework(Gabformer), (b) Multi-Gabor Self Attention (MGSA), (c) Gabor Filter (GAB), (d) Gated Feed-Forward Network (GFFN)}
	\label{fig:res}
\end{figure*}

The main contributions of our paper are summarized as follows:
\begin{itemize}
	\item We design a  multi-scale Gabor texture extraction filter that instructs the network to focus on high-frequency information, allowing the network to acquire richer contextual semantic information that helps recover image structure and texture details.
	\item We introduce a unique gating module (GFFN) to filter the information, i.e., the unimportant high-frequency information extracted by the Gabor filter is suppressed and only valid information is allowed to pass through the network.
	\item We have conducted extensive experiments on commonly used benchmark datasets to demonstrate the effectiveness and generalizability of the method. The experiments show that Gabformer can achieve excellent results in a wide range of rain scenes.
\end{itemize}
\section{PROPOSED METHOD }
\label{sec:format}
\subsection{Overall pipeline}
\label{ssec:subhead}
 Specifically, given a rainy image ${I_{in}} \in {\mathbb{R}^{M\times 3}}$, it is first processed through a generalized convolutional layer along the RGB channel to form low-level feature embeddings ${X_t} \in {\mathbb{R}^{M \times C}}$, where ${M} = {{H \times W}}$, \textit{H} and \textit{W} are the height and width of the feature map \textit{M},  \textit{C} denotes the number of channels. Feature processing is performed  step-by-step by an encoder-decoder model with a 4-level symmetric structure, which each level consists of multiple transformer blocks. In order to extract feature representations layer by layer, spatial size reduction of the feature map and channel expansion are achieved by stepwise encoding and decoding operations in the adjacent layers of the encoder. Pixel-unshuffle and pixel-shuffle operations\cite{14} are used in the downsampling and upsampling process of the features. In addition, use skip connections\cite{15} to introduce more contextual information and facilitate the flow of information. Finally, the refined features are processed through a convolutional layer and the resulting residual image is summed with the original input image to get the derained image, and the overall pipeline of our Gabformer architecture is shown in Fig 2. Formally, given the input feature ${X_{N - 1}}$ at the (N-1)-th block, the definition of the encoding process for Gabformer can be formulated as follows:
\begin{gather}
	{X_N} = {{\tilde X}_N} + GFFN(l({{\tilde X}_N})),\nonumber\\
	{{\tilde X}_N} = {X_{N - 1}} + MGSA(l({X_{N - 1}})),
\end{gather}
where $l( \cdot )$ is the layer normalization, ${\tilde X_N}$ and ${ X_N}$ are output features of the MGSA and GFFN modules.
% Table generated by Excel2LaTeX from sheet 'Data'
\begin{table*}[htbp]
	\centering
	\caption{Comparison of quantitative results from rain streak datasets, \textbf{bold} indicates best results.}
	\tabcolsep=0.27cm
	\small
	\begin{tabular}{p{4.165em}cccccccccccc}
	 \toprule
		Datasets &         &  \multicolumn{2}{c}{Rain200L\cite{2017Rain200L/H}} &       & \multicolumn{2}{c}{Rain200H\cite{2017Rain200L/H}} &       & \multicolumn{2}{c}{DDN-Data\cite{2017DDN-Data}} &       & \multicolumn{2}{c}{DID-Data\cite{2018DID-Data}} \bigstrut[t]\\
		Methods &       & \multicolumn{1}{c}{PSNR} & \multicolumn{1}{c}{SSIM} &       & \multicolumn{1}{c}{PSNR} & \multicolumn{1}{c}{SSIM} &       & \multicolumn{1}{c}{PSNR} & \multicolumn{1}{c}{SSIM} &       & \multicolumn{1}{c}{PSNR} & \multicolumn{1}{c}{SSIM} \bigstrut[b]\\
	 \toprule
		DSC\cite{DSC}   &       & 27.16 & 0.8663 &       & 14.73 & 0.3815 &       & 27.31 & 0.8373 &       & 24.24 & 0.8279 \bigstrut[t]\\
		GMM\cite{GMM}   &       & 28.66 & 0.8652 &       & 14.50  & 0.4164 &       & 27.55 & 0.8479 &       & 25.81 & 0.8344 \\
		MSPFN\cite{MSPFN} &       & 38.58 & 0.9827 &       & 29.36 & 0.9034 &       & 32.99 & 0.9333 &       & 33.72 & 0.9550 \\
		PReNet\cite{PReNet} &       & 37.80  & 0.9814 &       & 29.04 & 0.8991 &       & 32.60  & 0.9459 &       & 33.17 & 0.9481 \\
		RCDNet\cite{RCDNet} &       & 39.17 & 0.9885 &       & 30.24 & 0.9048 &       & 33.04 & 0.9472 &       & 34.08 & 0.9532 \\
		MPRNet\cite{MPRNet} &       & 39.47 & 0.9825 &       & 30.67 & 0.9110 &       & 33.10  & 0.9347 &       & 33.99 & 0.9590 \\
		CCN\cite{CCN}  &       & 38.26 & 0.9812 &       & 29.99 & 0.9138 &       & 32.67 & 0.9252 &       & 32.13 & 0.9238 \\
		SPDNet\cite{SPDNet} &       & 40.50  & 0.9875 &       & 31.28 & 0.9207 &       & 33.15 & 0.9457 &       & 34.57 & 0.9560 \\
		SwinIR\cite{SwinIR} &       & 40.61 & 0.9871 &       & 31.76 & 0.9151 &       & 33.16 & 0.9312 &       & 34.07 & 0.9313 \\
		Restormer\cite{zamir2022restormer} &       & 40.99 & 0.9890 &       & 32.00    & 0.9329 &       & 34.20  & 0.9571 &       & 35.29 & 0.9641 \\
		IDT\cite{2022IDT}   &       & 40.74 & 0.9884 &       & 32.10  & 0.9344 &       & 33.84 & 0.9549 &       & 34.89 & 0.9623 \\
		DRSformer\cite{DRSformer} &       & 41.23 & 0.9894 &       & \textbf{32.18} & \textbf{0.9330} &       & 34.36 & 0.9590 &       & 35.38 & \textbf{0.9647} \\
		Ours  &       & \textbf{41.71} & \textbf{0.9900} &       & 31.80  & 0.9280 &       & \textbf{34.45} & \textbf{0.9607} &       & \textbf{35.38} & 0.9629 \bigstrut[b]\\
		 \toprule
	\end{tabular}%
	\label{tab:addlabel}%
\end{table*}%
% Table generated by Excel2LaTeX from sheet 'data2'
\begin{table*}[htbp]
	\centering
	\caption{Comparison of quantitative results from a raindrop dataset, \textbf{bold} indicates best results.}
    \small
    	\tabcolsep=0.18cm
	\begin{tabular}{p{3.945em}cccccccccccccc}
		\hline
		Methods &       & \multicolumn{1}{c}{Pix2Pix\cite{Pix2Pix}} &       & \multicolumn{1}{c}{CMFNet\cite{CMFNet}} &       & \multicolumn{1}{c}{AttentGAN\cite{AttentGAN}} &       & \multicolumn{1}{c}{CCN\cite{CCN}} &       & \multicolumn{1}{c}{Quan’s\cite{Quan}} &       & \multicolumn{1}{c}{IDT\cite{2022IDT}} &       & \multicolumn{1}{c}{Ours} \bigstrut\\
		\hline
		PSNR  &       & 27.20 &       & 31.49  &       & 31.59 &       & 31.34 &       & 31.37 &       & 31.63 &       & \textbf{32.01} \bigstrut[t]\\
		SSIM  &       & 0.8359 &       & 0.9330 &       & 0.917 &       & 0.9293 &       & 0.9183 &       & 0.9361 &       &\textbf{0.9493}   \bigstrut[b]\\
		\hline
	\end{tabular}%
	\label{tab:addlabel}%
\end{table*}%
\subsection{Multi-Gabor Self Attention}
\label{ssec:subhead}
The Gabor filter is a spatial-frequency filter widely employed in the fields of image processing and computer vision, which uniquely combines Gaussian distribution and sinusoidal components to give it both spatial and frequency domain localization properties\cite{16}. This filter responds to a variety of structures in an image at multiple scales and orientations, and exhibits excellent performance in texture and edge extraction, it is described as the follows:
\begin{gather}
	G(x,y;\lambda ,\theta ,\psi ,\sigma ,\gamma ) = \exp ( - \frac{{{{x'}^2} + {\gamma ^2}{{y'}^2}}}{{2{\sigma ^2}}})\cos (2\pi \frac{{x'}}{\lambda } + \psi ),\nonumber\\
	x' = x\cos \theta  + y\sin \theta ,\nonumber\\
	y' =  - x\sin \theta  + y\cos \theta {\rm{,}}
\end{gather}
\begin{figure*}[htb]
	\begin{minipage}[b]{1.0\linewidth}
		\centering
		\centerline{\includegraphics[width=0.95\textwidth]{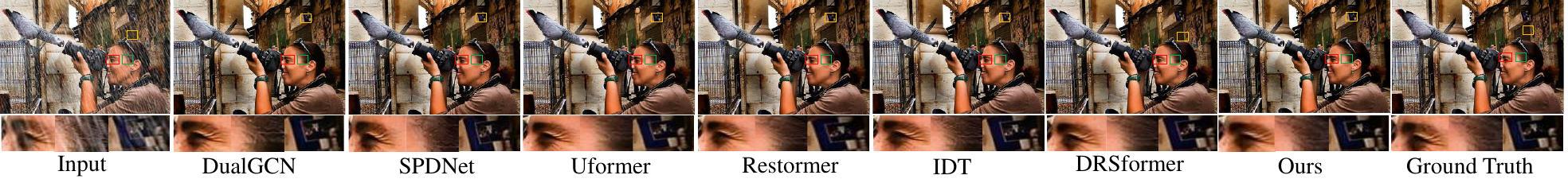}}
		%  \vspace{2.0cm}
	\end{minipage}
	\caption{Comparison of rain removal visualization on the Rain200L dataset, zoom in for clearer view of effectiveness.}
	\label{fig:res}
\end{figure*}
\begin{figure*}[htb]
	\begin{minipage}[b]{1.0\linewidth}
		\centerline{\includegraphics[width=0.95\textwidth]{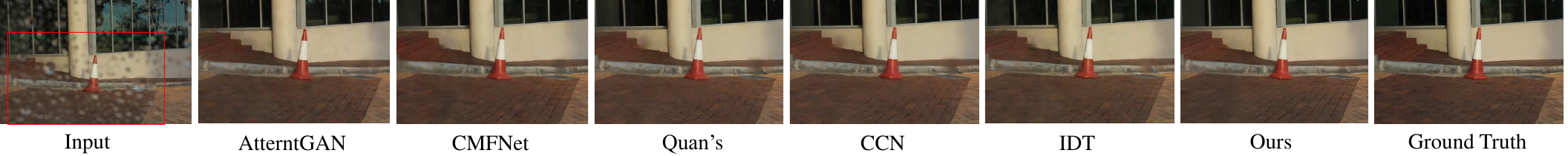}}
		%  \vspace{2.0cm}
	\end{minipage}
	\caption{Comparison of rain removal visualization on the AGAN-Data dataset, zoom in for clearer view of effectiveness.}
	\label{fig:res}
\end{figure*}
where \textit{x} and \textit{y} are the horizontal and vertical pixel coordinates , $\lambda$ denotes the filter wavelength, $\theta$ is the main direction of the filter, $\psi$ denotes the phase shift, $\sigma$ is the standard deviation of the Gaussian distribution, $\gamma$ denotes the spatial ellipticity of the filter, ${x'}$ and ${y'}$ are the coordinates obtained by rotating \textit{x} and \textit{y} in the original coordinate system. The larger the wavelength, the wider the range of image structures perceived by the filter, so we used four filters with different wavelengths to enhance adaptability to various scale structures within the image and deepen the overall understanding of the image content. By maintaining fixed values for phase offset, standard deviation, and spatial ellipticity, we ensured that the filter has a consistent texture response in a given direction and frequency. However, image edges often have significant gradients or slopes in one direction due to brightness variations. If only the wavelength is varied without changing the direction, the Gabor filter will  be able to detect edges and extract texture in a given direction, which may limit the specificity in different directions. Therefore, we used four directions at each wavelength based on the periodicity of the sine and cosine functions.Superimposing of information from multiple directions and wavelengths gives the model the flexibility to capture edge and texture information in all directions, thus improving the comprehensiveness and robustness of the image analysis. 

Inspired by \cite{zamir2022restormer}, We use cross-channel attention instead of traditional self-attention to save computational complexity. Given a layer normalized input tensor  ${X} \in {\mathbb{R}^{M \times C}}$,  we use 1×1 convolution for channel-wise context integration, followed by 3×3 depth-wise convolution to capture the spatial context within the channel to obtain the key and value. For query  extraction, we use 1 × 1 convolution channel context integration, followed by a Gabor filter with four different wavelengths, each with four orientations, to extract information within the channel dimension, which is given as:
\begin{gather}
Q = {f_{1 \times 1}}(\sum\limits_{i = 1}^4 {\sum\limits_{j = 1}^4 {Gab(X,G({\lambda _i},{\theta _j})))} } ,
\end{gather}
 where \textit{Q}, \textit{K} and \textit{V} are of dimension ${\mathbb{R}^{HW \times C}}$,  ${f_{1 \times 1}}( \cdot )$ is a 1×1 point-by-point convolution, $G( \cdot )$ stands for the gabor filter function, ${\lambda _i}$ and ${\theta _j}$ denote the i-th wavelength and the j-th direction,  $Gab( \cdot )$ is the filtering operation. Calculating Attention can be expressed as follows:
 \setlength{\abovedisplayskip}{8pt}
 \setlength{\belowdisplayskip}{8pt}
\begin{gather}
	Att(Q,K,V) = \varphi (Q,K)V,\nonumber\\
	{\rm{ }}\varphi (Q,K) = softmax (\frac{{{K^T}Q}}{\xi }),
\end{gather}
where ${\xi }$  is a scale parameter that can be learned to control the size of the attention map. By performing the attentional computation in the channel dimension, we change the computational complexity from $o(M \times M)$ to $o(C \times C)$ ,which is reduced due to $M \gg C$. Similarly, we divide the channels into multiple heads and train attention mappings for each head independently in parallel to improve the model's ability to capture complex relationships. Eventually, the output ${\tilde X}$ of the attention can be shown as:
\begin{gather}
 \tilde X = {f_{1 \times 1}}(Att(Q,K,V)) + X
\end{gather}
\subsection{Gating Feed-Forward Network}
\label{ssec:subhead}
Although we effectively reduce the computational complexity of the model by introducing cross-channel attention, this is accompanied by the use of a Gabor filter to extract texture features, which introduces additional computational overhead. In order to improve information transfer efficiency while reducing computational costs, we introduce a gating module\cite{zamir2022restormer} to replace the FFN. Given an input feature${\tilde X} \in {\mathbb{R}^{M \times C}}$, subsequent to layer normalization, we use a 1×1 convolution to expand the channel by a ratio of ${\varepsilon} $, then it is fed into two parallel paths, both paths are convolved with 3×3 deep-wise convolution to extract the local information, and one of them is activated with a \textit{GELU} for gating action. The final output is made by superimposing the features of the two paths, the whole process of fusing features is represented as
\begin{gather}
	\hat X = {d_{3 \times 3}}({f_{1 \times 1}}(l(\tilde X))\nonumber\\
	X = \tilde X + \hat X \odot GELU(\hat X)
\end{gather}
where ${\tilde X }$ denotes the output of the Multi-Gabor Self Attention,  ${d_{3 \times 3}}( \cdot )$ is a 3×3 deep-wise convolution, ${\odot}$ is the element-wise multiplication,  and ${ X }$ represents the output of the GFFN.

\section{EXPERIMENTAL RESULTS}
\label{sec:pagestyle}
\subsection{Experimental Settings}
\label{ssec:subhead}
\textbf{datasets.} Our experiments include five public benchmark datasets. The rain streak datasets include Rain200L\cite{2017Rain200L/H}, Rain200H\cite{2017Rain200L/H}, DID-Data\cite{2018DID-Data} and DDN-Data\cite{2017DDN-Data}. In addition, we evaluated the performance of the model using a raindrop dataset, \emph{i.e}, AGAN-Data\cite{2018AGAN-Data}.
\newline
\textbf{Implementation Details.} We used PSNR and SSIM to quantitatively evaluate the performance of the method. In our model, 
$\left\{ {{N_1},{N_2},{N_3},{N_4}} \right\}$ in Fig. 2 are $\left\{ {4,6,6,8} \right\}$, respectively.  The initial learning rate is set to $3 \times {10^{ - 4}}$  and decreases using a cosine annealing scheme to 
$1 \times {10^{ - 6}}$after 92,000 iterations, followed by another 208,000 iterations of model training. In the Gabor filter, ${\sigma}$ is set to $2\pi $, ${\psi}$ is 0,  ${\gamma}$ is 0.5, 
$\left\{ {{\lambda _1},{\lambda _2},{\lambda _3},{\lambda _4}} \right\}$ and 
$\left\{ {{\theta _1},{\theta _2},{\theta _3},{\theta _4}} \right\}$ are 
$\left\{ {1.0,1.5,2.0,2.5} \right\}$ and 
$\left\{ {{{45}^ \circ },{{90}^ \circ },{{135}^ \circ },{{180}^ \circ }} \right\}$ , respectively. Table 3 shows the effect of using filters with different wavelengths and orientations on the rain removal effect of the image. From Table 3, the effect is significantly improved by superimposing multiple wavelengths and directions.
\begin{table}[htbp]
	\centering
	\caption{Parametric sensitivity analysis, we employed diverse Gabor filter with different wavelengths and orientations, comparing PSNR on the Rain200L dataset to evaluate feature extraction performance.}
	\tabcolsep=0.09cm\
	\small
	\begin{tabular}{ccccccccccc}
		\hline
		\multicolumn{1}{c}{Metric} &       & ${\theta _1}$     &       & ${\theta _2}$     &       & ${\theta _3}$     &       & ${\theta _4}$     &       & ${\theta _1,\theta _2,\theta _3,\theta _4}$ \bigstrut\\
		\hline
		${\lambda _1}$     &       & 41.42 &       & 41.30  &       & 41.13 &       & 41.37 &       & 41.06 \bigstrut[t]\\
		${\lambda _2}$     &       & 41.30  &       & 41.22 &       & 41.41 &       & 41.15 &       & 41.29 \\
		${\lambda _3}$     &       & 41.20  &       & 41.15 &       & 41.41 &       & 41.35 &       & 41.48 \\
		${\lambda _4}$     &       & 41.16 &       & 41.21 &       & 41.17 &       & 41.35 &       & 41.36 \\
		${\lambda _1,\lambda _2,\lambda _3,\lambda _4}$     &       & 41.51 &       & 41.63 &       & 41.48     &       & 41.35 &       & \textbf{41.71} \bigstrut[b]\\
		\hline
	\end{tabular}%
	\label{tab:addlabel}%
\end{table}%
\newline
\textbf{Quantitative Evaluation.} We compare our Gabformer with the state-of-the-art 
methods, including two prior-based models (DSC\cite{DSC} and GMM\cite{GMM}), as well as CNN-based approaches (including MSPFN\cite{MSPFN}, PReNet\cite{PReNet}, RCDNet\cite{RCDNet}, MPRNet\cite{MPRNet}, CCN\cite{CCN}, and SPDNet\cite{SPDNet}). In addition, we examine recent Transformer-based approaches, including Uformer\cite{Uformer}, SwinIR\cite{SwinIR}, Restormer
\cite{zamir2022restormer} , IDT\cite{2022IDT}, DRSformer\cite{DRSformer}. For raindrop removal, we compare CMFNet\cite{CMFNet}, Pix2Pix\cite{Pix2Pix}, Atten-tGAN\cite{AttentGAN}, Quan's network\cite{Quan}, CCN\cite{CCN} and IDT\cite{2022IDT}. Table 1 shows the results of comparing our method with the SoTA method for rain streak datasets, and Table 2 shows the comparison results for raindrop datasets. We can see that our method achieves much better performance on most datasets from Table 1 and Table 2. In addition, Fig. 3 and Fig. 4 show the rain removal effect of the images in the two datasets, which also demonstrate that our method has better recovery results.
\subsection{Ablation Study}
\label{ssec:subhead}
In this section, we evaluate the performance of the proposed Gabformer. Specifically, we evaluate the impact on queries of both adding the Gabor filter to extract texture features and feature extraction by depthwise separable convolution. Furthermore, validation is performed to compare the difference between the case where texture features are extracted by adding the Gabor filter only to ${Q_G}$ and the case where the gabor filter is applied to ${Q_G}$, ${K_G}$, ${V_G}$. In addition, it has been mentioned that using the relu function instead of the softmax  can generate sparse attention maps to effectively exclude irrelevant features\cite{sparse}, and we also conducted experiments. The experimental results are shown in Table 4. From table 4, our proposed Gabformer has the best performance.
% Table generated by Excel2LaTeX from sheet 'xiaorong'
\begin{table}[htbp]
	\centering
	\caption{Ablation study on the Rain200L dataset for image deraining with different network configurations}
		\tabcolsep=0.35cm
		
		\small
	\begin{tabular}{p{10em}cccc}
		\hline
		Network Configuration &       & \multicolumn{1}{c}{PSNR} &       & \multicolumn{1}{c}{SSIM} \bigstrut\\
		\hline
		${Q}, {K}, {V} + $\emph{GFFN} &       & 40.99 &       & 0.9890 \bigstrut[t]\\
		${Q_G},{K_G},{V_G} + $\emph{softmax+GFFN} &       & 41.28     &       & 0.9897 \\
		${Q_G},{K_G},{V_G}+$\emph{RELU+GFFN} &       & 39.17     &       & 0.9849 \\
		${Q_G},{K},{V}+$\emph{RELU+GFFN} &       & 41.00    &       & 0.9890 \\
		${Q_G},{K},{V}+ $\emph{softmax+FFN} &       & 39.28 &       & 0.9848 \bigstrut[b]\\
		${Q_G},{K},{V}+ $\emph{softmax+GFFN} &       & \textbf{41.73} &       & \textbf{0.9900} \bigstrut[b]\\
		\hline
	\end{tabular}%
	\label{tab:addlabel}%
\end{table}%
\subsection{Experimental Analysis}
\label{ssec:subhead}
In this study, we investigate the effectiveness of the multi-scale Gabor filter by varying the combination of wavelengths and directions, and the experimental results are shown in Table 3. The experimental results show that the filter with a mixture of multiple wavelengths and directions has better performance in image processing. This demonstrates the importance of multiscale information, as this mixture captures the features in the image more comprehensively. However, we also observe that certain specific combinations (e.g., [${\lambda _1}$ , ${\theta _1}$]) are able to produce more pronounced results at certain angles or scales. This may be due to the fact that the image has more significant gradients at these angles or scales, making the particular combination of filter more effective. Overall, the filter with a mixture of several wavelengths and directions is still the best choice.

We use the texture information extracted by Gabor as queries, and use depthwise separable convolution to encode spatially localized contextual information as keys and values, which allows the network to focus on more contextual semantic information in the image, improving the network's localization and globalization.

By using the Gabor filter, more high frequency information is given to the network, but not all of the high frequency information contributes to the image deraining, so we use a gated module (GFFN). The last two rows of the ablation experiments in Table 4 show the difference between the modules used. As can be seen in Table 4, the gated module effectively suppresses the less important information in the channel and allows the effective high frequency information to be delivered, resulting in excellent performance with a reduced number of network parameters.
\section{CONCLUSIONS}
\label{sec:pagestyle}
In this study, we propose a new Gabor-based transformer for single image deraining. Specifically, we design multi-scale and multi-directional combined the Gabor filter to extract image texture features and use them as queries for the attention mechanism to increase the network's attention to the detailed information of the image, which improves the deraining effect. In addition, by extracting texture features in the image channel dimension to reduce the computational complexity, our Gabformer can also be applied to high-resolution images. A large number of experiments surface that our proposed method works better than the SoTA method. However, the addition of Gabor filter introduces an additional number of parameters into the model. Our model has 34.4M parameters, which may be difficult to deploy on resource-limited devices, and we will effectively filter the useful high and low frequency information of the image in our model, using pruning or distillation to reduce the number of parameters of the model along with good performance.


\bibliographystyle{IEEEbib}

\bibliography{refs}

\end{document}